\documentclass[10pt,letterpaper]{article}
\usepackage[top=0.85in,left=1.75in,footskip=0.75in,marginparwidth=2in]{geometry}

\usepackage[utf8]{inputenc}

\usepackage{cite}

\usepackage{nameref,hyperref}

\usepackage[right]{lineno}

\usepackage{microtype}
\DisableLigatures[f]{encoding = *, family = * }

\raggedright
\usepackage{changepage}

\usepackage[aboveskip=1pt,labelfont=bf,labelsep=period,singlelinecheck=off]{caption}

\makeatletter
\renewcommand{\@biblabel}[1]{\quad#1.}
\makeatother

\usepackage{lastpage,fancyhdr,graphicx}
\usepackage{epstopdf}
\fancyheadoffset[L]{2.25in}
\fancyfootoffset[L]{2.25in}

\usepackage{color}

\definecolor{Gray}{gray}{.25}

\usepackage{graphicx}

\usepackage{sidecap}

\usepackage{wrapfig}
\usepackage[pscoord]{eso-pic}
\usepackage[fulladjust]{marginnote}
\reversemarginpar

\begin{document}
\vspace*{0.35in}

% title goes here:
\begin{flushleft}
{\Large
\textbf\newline{Assistive Chatbots for healthcare: a succinct review}
}
\newline
% authors go here:
\\
Basabdatta Sen Bhattacharya\textsuperscript{1,*},
Vibhav Sinai Pissurlenkar\textsuperscript{2}
% Author 3\textsuperscript{1},
% Author 4\textsuperscript{1},
% Author 5\textsuperscript{2},
% Author 6\textsuperscript{2},
% Author 7\textsuperscript{1,*}
\\
\bigskip
\bf{1} Computer Science and Information Systems, BITS Pilani Goa Campus, Goa, India
\\
\bf{2} Goa University, Goa, India
\\
\bigskip
* basabdattab@goa.bits-pilani.ac.in

\end{flushleft}

\section*{Abstract}
Artificial Intelligence (AI) for supporting healthcare services has never been more necessitated than by the recent global pandemic. Here, we review the state-of-the-art in AI-enabled Chatbots in healthcare proposed during the last 10 years (2013-2023). The focus on AI-enabled technology is because of its potential for enhancing the quality of human-machine interaction via Chatbots, reducing dependence on human-human interaction and saving man-hours. Our review indicates that there are a handful of (commercial) Chatbots that are being used for patient support, while there are others (non-commercial) that are in the clinical trial phases. However, there is a lack of trust on this technology regarding patient safety and data protection, as well as a lack of  wider awareness on its benefits among the healthcare workers and professionals. Also, patients have expressed dissatisfaction with Natural Language Processing (NLP) skills of the Chatbots in comparison to humans. Notwithstanding the recent introduction of ChatGPT that has raised the bar for the NLP technology, this Chatbot cannot be trusted with patient safety and medical ethics without thorough and rigorous checks to serve in the `narrow' domain of assistive healthcare. Our review suggests that to enable deployment and integration of AI-enabled Chatbots in public health services, the need of the hour is: to build technology that is simple and safe to use; to build confidence on the technology among: (a) the medical community by focussed training and development; (b) the patients and wider community through outreach.

% now start line numbers
% \linenumbers

% the * after section prevents numbering
\section{Introduction}
\label{sec:1}
Artificial Intelligence (AI) today has reached a stage where it is being perceived as a serious existential threat to the human kind (\url{https://www.bbc.co.uk/news/world-us-canada-65452940}), particularly with the recent introduction of the conversational agent ChatGPT (\url{https://openai.com/blog/chatgpt}). Belonging to the genre of Generative AI, the ChatGPT is essentially a deep neural network that has been trained over humongous sets of data over a period of around 7-8 years. This may be thought akin to training a baby from childhood to a stage where not only can they converse on the matters that they were trained on, but they can also tap into their past conversational experience to interact on matters that were not explicitly in their training framework. Indeed, conversational agents, commonly and succinctly referred to as Chatbots (Robots that can chat/converse/interact), was first proposed at the AI lab of Massachusetts Institute of Technology, USA~\cite{weizenbaum1966}, even if the idea of a thinking and interacting machine can be traced back to the famous imitation games proposed by Alan Turing~\cite{turing1950}. For healthcare, the emphasis on assistive AI has never been greater than around the time of the recent global pandemic. As the doctors, nurses and carers themselves became affected physically and mentally by the virus, and often forced to be off work due to long term effects of the disease, the burden on the medical services globally reached unprecedented levels. In particular, mental health has emerged as a major long term effect of the pandemic, even if it often tends to go unreported, perhaps due to inhibition of the patients to share their mental state and feelings, not to mention the attached stigma (\url{https://www.psychiatry.org/patients-families/stigma-and-discrimination}). One example of Nations trying to address this issue is the dedicated page for mental health of adolescents (\url{https://www.youngminds.org.uk/}) in the United Kingdom (UK), a section of the society that was one of the hardest hit by the pandemic. Furthermore, the National Health Services (NHS), UK have self-assessment pages for Mood of adolescents~\cite{mood} and Depression~\cite{depression}, which have a text-based user interface and have a Question and Answer (Q\&A) format. Even before the pandemic hit though, an independent review recommendation led by Dr. Eric Topol during 2018-19 (\url{https://topol.hee.nhs.uk/the-topol-review/}; on invitation by the UK Government) projects that around 80\% or more of patients and medical staff will use and/or benefit from using AI for ``Speech recognition and natural language processing (NLP)'' (see p 26 Fig 1); in this context, the report outlines how recent advances in the speech recognition (SR) technology ``now presents a valuable tool for clinical documentation – the benefit for the healthcare workforce to focus on patient interaction and care rather than the computer screen and keyboard is clear. This is likely to have a major impact in primary care, as well as outpatients and emergency departments in hospitals.'' The Topol review further outlines how an existing Chatbot is being used for triaging mental health patients in London, and the future possibilities (interested readers may refer to section 3.1.5 in the report). 

In this brief review, our objective is to highlight the Chatbots that were developed and designed specifically for healthcare (both mental and physical) applications during the last ten years i.e. between 2013~--~2023. In addition, we focus on Chatbots that (claim to) have used AI as a technology in developing the Chatbots. Here, we note that rule-based algorithms such as decision trees very much falls under the purview of AI as well as Machine Learning (ML), the latter itself being a subset of the former. Our reference to the term `Chatbot' in this article refers to all conversational agents that use AI (rule-based \& other ML) algorithms and can communicate with the user autonomously and require minimal support from human-in-the-loop.

In our literature survey, we have come across several Chatbots, both for mental and physical health, that are led by not-for-profit agencies and Institutions (including Higher Education, Medical, and Research Institutes and Organisations). A significant factor in the testing and deployment of Chatbots is their acceptance by the medical community as assistive tools. A survey and analysis study on a cohort of 100 professionals and experts in mental health~\cite{sweeney2021} report that 23\% of the participants thought that Chatbots were unimportant, and only 11\% thought that they were very important; the remaining thought they were somewhat important. Furthermore, most respondents were not aware of available Chatbots, although 80\% of them thought they were likely to prescribe assistive Chatbots within the next five years. The other major inhibitive factor may be attributed to a perception that over time, Chatbots may replace or reduce the number of medics in the profession. In this scenario, the recommendations for education and training to inform and familiarise the medical community with AI technology seems apt and timely. At the same time,~\cite{meadows2020} raise the important issue of assessment of the Chatbots in terms of their efficacy in patient recovery. On this, the term `recovery' is brought to the fore and the need for a consensus on the definitive (set of) symptom(s) that would define recovery in specific psychiatric conditions is emphasised. Here, we note the importance of evaluating the `Chatbot as an evaluator' using appropriate metrics.

In addition to non-commercial Chatbots, we have come across a few commercial ones that can be accessed at a cost to the individual or employer, of which we found three organisations that have dedicated Chatbots designed for mental health consultancy and education. We note that this is not by any means an exhaustive list of available commercial applications; rather a list selected based on their accessibility and user ratings on the web. Furthermore, and to the best of our knowledge, we did not come across a commercial deployment of Chatbots for physical health consultancy and treatment; thus, all the commercially available Chatbots that we have discussed here are for mental well-being. The greater emphasis on digital intervention for mental health may be attributed to the flagging of several psychiatric and psychological illnesses as a global health issue by the  World Health Organisation, leading to an increased funding for research on viable options for treatment, therapy and outreach. In a recommendation for research on mental health in the USA, conversational agents (i.e.\ Chatbots) as Behavioural Intervention Technologies (BITs)~\cite{miner2016} is mentioned as a viable pathway. However, the main hurdle at the time was the NLP interface that proposed the Relational Frame Theory (RFT) (``which posits that humans use linguistic frames to understand the world around them, and subsequently solve problems.'') as a means to design conversational agents that can be deployed for assessment of mental health patients. The emphasis was on building conversational interface that are informed by standard practices in psychological therapy.

Implementation of software for Chatbots is mainly via two modes: (a) Stand-alone computers or laptops; here, the conversational agent software either needs to be installed on the machine and executed locally, or is made available on a cloud server via a web browser. (b) Smart phones and tablets: here, the conversational agents are software applications (`Apps') that execute on the Google Android or Apple iOS operating systems (OS), available to install on the mobile devices from the Google Play and Apple stores respectively. Some of these Apps also run on the Microsoft Windows OS and thus can be accessed on stand-alone computers and laptops. Chatbots that are available via web browsers can also be accessed from these mobile devices. 

Overall, while several Chatbots are available (commercially) as well as being developed currently, a lot needs to be done in terms of rigorous testing and assessment to build confidence in the technology and integrate them with existing public health services. The relevance of Chatbots for medical intervention and support is made evident by a plethora of review articles that were published within a short span of 7-8 years; we make a brief overview of selected existing review articles in Sec.~\ref{sec:2}. In Sec.~\ref{sec:3} we discuss Chatbots that are developed for mental healthcare; Chatbots that are led by Academic and Medical Research Institutes and Organisations and are in several pre-deployment phases are presented in Sec.~\ref{sec:31}; Chatbots that are deployed commercially  are discussed in Sec.~\ref{sec:32}. In Sec.~\ref{sec:4}, we discuss the Chatbots for physical healthcare and wellbeing, all of which are in the research and development phases. We discuss our findings from this review and conclude the paper in Sec.~\ref{sec:5}.
\section{Existing literature surveys}
\label{sec:2}
Four researches over a period of 2 years (2017 -- 2019) capture well the progress of conversational agents in healthcare, which we present here (\cite{laranjo2018, provoost2017, alrazaq2019, bendig2019}). We have observed some differences in terminology regarding the different phases of the Chatbot life-cycle. Thus, while reading the papers, we classified the phases of the Chatbot life-cycle broadly into 5 categories (\url{https://chatbotsmagazine.com/the-bot-lifecycle-1ff357430db7}) viz.\ (i) research and design (ii) software development and implementation (iii) pilot testing (iv) deployment (v) post-deployment testing and evaluation. The post-deployment maintenance phases are irrelevant for this review as we did not find any non-commercial Chatbot currently in healthcare that are in this stage; for the commercial Chatbots, we do not have enough information to be able to comment on this. For the benefit of the reader, we would like to note here that testing of Chatbots in healthcare needs `clinical trials' to ensure patient safety and efficacy of treatment. Randomised Control Trials (RCT)~\cite{rct2018} is considered as the gold standard in clinical trials. Many of the Chatbots for mental health that are discussed here underwent RCT for pilot testing; PHQ (Public Health Questionnaire) and GAD (General Anxiety Disorder) metrics are used for quantitative evaluation of Depression and Anxiety levels in patients (for example, see \url{https://talk2gether.nhs.uk/quick-self-assessment/}) participating in clinical trials for piloting/evaluating Chatbot efficacy.

The systematic review by~\cite{laranjo2018} helped us to understand the breadth of the work done in the area of our interest i.e.\ Chatbots in healthcare. In fact, the authors mention that this is the first instance of a review on conversational agents in healthcare where these agents can also communicate in modes other than in natural language e.g. facial expressions, body movements. After a thorough process of reviewing existing research reports, the authors focus on 14 reports on conversational agents that are out of the development phases and passed the pilot phase, with the strict condition that the agents were tested on human users (those that were not, are excluded from their review). We note that the authors mention chatbots and Embodied conversational agents (ECA) only as subsets of the umbrella term `conversational agents'; chatbots are those that has the ``ability to engage in small talk and casual conversation"; ECAs involve a computer generated `avataar' and provide an illusion of a `face-to-face' communication with a human. Of the 14 conversational agents reviewed, 5 were ECAs and 2 were chatbots. (Readers may note that to avoid terminology clash, we have used the upper-case for `c' in `Chatbot', a term that we have used to refer to all conversational agents in our review, including ECAs). The authors specify six fundamental attributes as design choices for conversational agents as well as technical measures for evaluating their efficacy (See their Boxes 1 and 2 respectively). Overall, the review findings indicates an alarming gap in consensus among research groups and institutions on testing and evaluating methodologies for conversational agents. Furthermore, the review recommends incorporating patient safety as an evaluative metric for conversational agents. 

In a scoping review,~\cite{provoost2017} focus on ECAs for mental health only, and identify 49 studies in the field. The higher count than that reported in~\cite{laranjo2018} is because this study includes ECAs that are also in the development or pilot phases. The communication with the ECAs were primarily through speech ($\approx 37\%$), facial expressions ($\approx 31\%$) and hand and body gestures ($\approx 29\%$). Furthermore, $\approx 53\%$ of the ECAs targetted Autism Spectrum disorder (ASD), and around $\approx 20\%$ targetted depression. And yet, only 7 of the 49 ECAs reported here moved beyond the pilot phase. Out of these, 6 were in the evaluation phase (2 each in ASD and depression, 1 each in substance use and psychotic disorder), and only 1 (depression) reached the implementation (i.e.\ deployment) phase. For ASD, ECAs were mainly used for education and skills training purposes where the ECA is posited as the trainer or coach. For depression too, the ECAs were used as a virtual coach or healthcare personnel; interventions were through counselling, skills training and cognitive behavioural therapy (CBT). We did not find a specific reference to the ECA in depression that is counted as `implemented'. In agreement with that reported in~\cite{laranjo2018}, the evaluation of the ECA `performance' in psychotherapy for each of the 7 disorders reviewed here varies across studies both in terms of the evaluation criteria and methodologies. The author recommend low tech, robust, simple technology for ECAs, which will facilitate fast progression from the development and pilot phases to deployment for `routine clinical practice’ in a safe and effective manner. 

Another scoping review~\cite{alrazaq2019}, focussing on the Chatbots for mental healthcare (only), report a mind boggling 15000 available mobile healthcare apps, out of which 29\% were for mental healthcare. Of these, the majority used decision tree algorithms for developing the Chatbots; 2 research papers proposed supervised learning and reinforcement learning (RL) algorithms respectively; another 2 claimed AI-based rules for their Chatbots, but failed to specify the exact methodologies and network architecture. The interest in the research community seems recent, starting around 2005; around 50\% of the research were during the period 2016 - 2019. More than 17 countries worldwide are involved in research in this area, out of which only 4 developing countries are involved. The authors seem to make a subtle hint that developing countries could benefit more from this technology compared to developed countries. On this, we note that one way of building acceptance of Chatbots in developing countries, suggested by~\cite{singh2019} while warning on a looming ``mental health epidemic'' in India, could be by integration of the technology with indigenous language and culture.

Around the same time,~\cite{bendig2019} made a scoping review with a focus on the efficacy of available Chatbots in mental healthcare (only). The authors note that Chatbots can make valuable contributions to the psychotherapy procedures where contact time between patient and the practitioner is not essential, for example psycho-education, goal-setting conversations, therapeutic writing. Furthermore, an empathetic Chatbot may be more accepted by the patients and practitioners. They recommend quality checks on patient safety when working with Chatbots, as well as a need for evolving psychotherapy database of the existing bots.

The systematic review by~\cite{laranjo2018} indicates that there is an equal number of conversational agents in physical (8) and mental health (8) (and one chatbot reported as being piloted for both physical and mental health; see their Table 3) that are developed and piloted on human users. Existing commercial Chatbots were not discussed, perhaps because they did not meet the selection criteria for the review. Indeed, a few commercial Chatbots are mentioned in~\cite{bendig2019}, with a note that these are not `empirically tested'. 

Our review adds value to the above-mentioned literature survey in three aspects: First, our interest is in Chatbots that incorporate AI as a technology (for example Decision Trees, NLP, DL) for patient support, therapy, as well as educational and coaching activities. Second, wherever available, we are interested in the specific AI technique used to design and develop the Chatbot. Last but not the least, we report on the current status of the Chatbots that we have discussed here. 

For this brief review, we did not make a comprehensive coverage of all available Chatbots in healthcare. Rather, we have taken cues from the bibliography of the existing reviews, a strategy termed as `backward snowballing'. Next, we have used the Google Search Engine (GSE) with search keywords consisting of those from the specific bibliography. Of the several such references, we have focussed only on Chatbots that used AI technology for patient treatment and therapy. In addition, for App-based Chatbots, we have considered only two that ranked high on user ratings, at the same time had `AI' mentioned in their description. 
\section{Chatbots for mental health and wellbeing}
\label{sec:3}
In Sec.~\ref{sec:31}, we discuss selected Chatbots in mental healthcare that are being developed by leading Academic, Research and Medical Institutions and are not yet deployed for wider public health services. We have come across several commercial Chatbots that highlight their use of AI technology. However, and understandably so, we did not find the specifics of the AI technique used to design and develop these Chatbots. In Sec.~\ref{sec:32} we discuss selected commercially available Chatbots that scored well on user/customer review ratings. 
\subsection{Non-commercial Chatbots}
\label{sec:31}
The most trialed non-commercial Chatbot currently is `\textit{Woebot}'~\cite{fitzpatrick2017}. The conversational agent was built using Decision Tree and appropriate NLP algorithms and needs to be installed as a software in a stand-alone computer. All conversational responses are stored in a database. The specific area of application is Cognitive Behavioural Therapy (CBT) for anxiety and mood disorder targeting young adults that are college going students. During an evaluation exercise, participants were tutored on `how to use' Woebot. There were several categories on which Woebot could provide therapy. PHQ-9 score (``a 9-item self report questionnaire that assesses the frequency and severity of depressive symptomatology within the previous 2 weeks."~\cite{fitzpatrick2017}) was used to assess the state of the participants. Results showed that there was a significant improvement in the mean PHQ-9 score compared to controls in the mental condition of participants with 2 weeks of therapy. This was similar to other mobile app based interventions~\cite{watts2013,kauer2012,burns2011}, however, the time taken for symptom improvement was in a quarter of the time (2:8 weeks). The only other study that is reported to have a significant effect within 2 weeks is a software application for treating borderline personality disorder. Readers may note that the proposed software applications viz.\ Mobilyz~\cite{burns2011}, Mobiletype~\cite{kauer2012} were for dealing particularly with depressive disorders. A comparison between mobile-based and stand-alone-computer-based intervention for depression showed that using mobile did not in any way affect the efficacy of the programme~\cite{watts2013}. However, these software applications were not Chatbots, rather, they were strategies to use technology for coaching and therapy under the supervision of medical practitioners. However, and more recently,~\cite{gutu2021} studied the efficacy of Woebot on a cohort of 95 adults (age range 19 -- 43 years) all of who were suffering from either depression or anxiety. This research finds that improvement after a 2-week program via Woebot was no different than that obtained via simple techniques such as ``regular emails sent to [sic] people". Furthermore, they reported a higher attrition rate in the cohort that were taking the Woebot-based therapy, which the authors speculate to be linked to Chatbots ``still grappling with limitations in terms of content". In making this remark, the authors are not just pointing to Woebot, rather to all such that are available to use.

In the aftermath of the Covid pandemic, a web-based conversational Chatbot `Mental health Intelligent information Resource Assistant' (MIRA) was proposed~\cite{noble2022} in Canada. The authors claim that what sets MIRA apart in current times is that it supports ``a wide range of mental health disorders”. The application itself is built on Rasa, an open source conversational AI platform that is described as `human, smart, and intelligent'(\url{https://rasa.com/blog/the-humans-behind-the-bots-andrea-kelemen/}). The questionnaire was created by a trained psychiatrist, which was then improved over several iterations by a multidisciplinary, multicultural team of researchers; the target beneficiaries of this Chatbot were the healthcare workers and their families in the Alberta and Nova Scota provinces of Canada. Thus, MIRA is not meant for providing healthcare solutions and services to patients; rather, the main purpose is to provide information to users. Since the publication of this article, MIRA seems to have developed further; the latest updates are available here:\url{https://www.mymira.ca/index.html}. 

In a series of researches under the Northern Periphery and Arctic Programme (NPAP; ~\url{https://chatpal.interreg-npa.eu/}),~\cite{boyd2022} proposed ChatPal, an Android- and ios App-based Chatbot primarily targetting the mental health and well-being of the rural population. Chatpal was tested on a cohort of around 348 citizens in the age range of $18-73$ years and with various professions such as students, teachers, assistant nurses, unemployed. Of these, 18\% thought ChatPal was `very useful’, 53\% thought it was ‘somewhat useful’, 10\% rated it as neutral; the remaining did not report the bot to be useful at all. When the cohort were asked how much they believed the Chatbot changed their mental wellbeing, 83\% reported either slight improvement or none at all. Overall, the authors did not notice any significant improvement in mental wellbeing using ChatPal; however, they propose blending the tool in normal mental health services such as to support the practitioners. We note that in an earlier report where these authors assessed the demand for a Chatbot from participants of a workshop~\cite{potts2021}, it seemed that participants liked the idea of the Chatbot regularly checking-in with the user, asking questions about emotional state or mood and tracking this over time. For repeated use of the Chatbot, participants felt that reflecting on previous conversations would be beneficial. Many also thought that the Chatbot should provide a space to share thoughts and feelings but also provide information. However, when indeed the Chatbot was proposed, and then tested on a different cohort, the outcome was certainly not as positive as during the assessment task. This indicates the complexity and variability in design and implementation of Chatbots. ChatPal is an ongoing development project, where the latest report is on implementation of event logging facilities to record and analyse the usage of the bot~\cite{potts2023}.

Another Chatbot `\textit{iHelpr}’~\cite{cameron2019, cameron2017} is a text-based interactive Chatbot intended to provide mental health support in the workplace. The iHelpr is a web-based self-assessment tool and is reported to be available for six well-being indicators viz.\ stress, anxiety, depression, sleep and self esteem. A high score implied that the employee needed help, and were directed to a helpline number and an emergency contact. The Chatbot is developed using the bot development framework by Microsoft’s Cognitive Services, ``an Application Programming Interface (API) that can process natural language, enable a Chatbot to recognise speech, and image-processing technology''~\cite{cameron2018a}. Furthermore, through knowledge engineering in the form of extensive interactions with domain experts (clinical psychologists), a database was created that stores coping strategies and questionnaire scores. The iHelpr is embedded within an online self help portal called Inspire Support Hub, and refers to the database for providing its responses and scores to the user.

An interesting application developed and trialed at the Ulm University, Germany~\cite{bendig2021}, is a script-based conversational agent, which they call \textit{S}oftware agent providing an \textit{I}ntervention for \textit{S}elf-help to \textit{U}plift psychological wellbeing (“SISU”). SISU is indeed a Chatbot, but it is different in providing support for therapeutic writing. Writing one's negative life events, past or ongoing, is well known to be therapeutic and is a recommended strategy for healing and upliftment of mood. In the trial conducted with SISU and reported to be the first of it's kind, SISU guided participants to write everyday for a set time on some negative life event of their choice. Thereafter, based on principles of Acceptance and Commitment Therapy (ACT) and narrative psychology, SISU guided each participant to reduce negative thoughts and feelings. The application is intended to be accessible via mobile and is linked to the text-based services of the mobile network. However, we did not find any reference to use of AI tools for designing SISU (see~\url{https://oparu.uni-ulm.de/xmlui/handle/123456789/45489}).

A generative Chatbot was proposed by~\cite{more2021} as a communication interface to relieve stress. The model is trained using two LSTM networks and a SVM classifier is used to detect the mood and then provide an appropriate response. An attention mechanism is embedded in their algorithm, where the bot is shown as responding in a sensible manner in a simple conversational exchange. However, the paper did not elucidate on deployment or future plans along such lines.

In summary, among the Chatbots that are being developed and disseminated by not-for-profit organisations, MIRA and Chatpal are in several stages of clinical trials, and seem to have a definite plan for deployment in the near future. While Woebot has been clinically trialled by several research groups, we did not find any resource indicating plans for integration with health services for the wider community. Along similar lines, we haven't seen instances of plans for piloting and deployment for the remaining Chatbots that we discussed here.
\subsection{Commercially available Chatbots}
\label{sec:32}
\begin{itemize}
    \item \textbf{Tess} is a web-based chatbot devised by X2AI Inc. (\url{https://tess.x2ai.com/}) with access interface via SMS (on mobile) and Facebook Messenger application. In a technical report (where authors are affiliated to X2AI),~\cite{joerin2019} provides a synopsis on the three stages of Tess's design and implementation. The technology is based on machine learning algorithms integrated with psycho-educational concepts and is said to be developed in conjunction and collaboration with trained mental health professionals. They refer to Tess as a ``psychological artificial intelligence service'', designed to provide much needed support to healthcare professionals, primarily targetting the services in United States (US) and Canada, whose mental unwellness often goes unreported. In the second phase, extensive effort was put in terms of scripting Tess's database by skilled professionals. The third phase is reported as a work in progress (during 2019) where Tess was being prepared to be tested on patient groups.
    
    Today, on their website, X2AI claims that ``Tess has been clinically validated to help people feel better: Chatting with Tess was found to reduce symptoms of depression (-13\%) and anxiety (-18\%).'' Indeed,~\cite{fulmer2018} performed a randomised control trial to observe the efficacy of Tess on University students in the USA. They reported a significant decrease in both depression and anxiety levels after 2--4 weeks, measured on PH-9 and GAD-7 scales respectively and using multivariate analysis of covariance (manova). The average age of the 74 participants was 22.9, where 70\% were female, and 51\% and 41\% were of Asian and White respectively. The control group in the study were given access to an `e-book on depression among college students' (\url{https://www.nimh.nih.gov/health/publications/depression-listing}), as opposed to communicating with Tess like the test groups. Also, there were two test groups; one that received daily message from Tess on varying contents over a period of 2 weeks, and the other that received biweekly message from Tess but on a specific content over a period of 4 weeks. Both test groups showed significant decrease in symptoms upon interacting with Tess. However, user satisfaction survey received a mixed response with many participants reporting Tess getting confused and misunderstanding when receiving unexpected user input. 
    
    Another study~\cite{dosovitsky2020}, while looking at efficacy of Tess to deal with depressive symptoms, was unique in providing an insight into the chatbot's framework, i.e.\ how would the users be taken through the contents implementing BITs. The paper explains the `module' format of Tess and studies the chatbot by observing how 354 users interacted with Tess's depression modules. Overall, their study recommends further work that needs to be done to make current Chatbots to be user friendly as well as trustworthy in context to the job that it is entrusted with. 
    
    Similar recommendations were also provided in a commentary~\cite{thompson2019} on a feasibility study addressing pediatric obesity~\cite{stephens2019}, a public health matter of growing concern in the USA, that used Tess for CBT of obese children.~\cite{thompson2019} raises some very apt and timely concerns, the most important of which are: first, there is still not enough labelled data available for Tess to perform reliable therapy; second, not enough safety and privacy related concerns are addressed, especially where children are concerned; third, the cohort on which Tess was tested was from 9-18 years that seemed generic rather than targetted and personalised treatment, considering children grow at faster and variable rates specifically during this age range.

    From this review, our assessment is that Tess looks promising in terms of coaching- and conversational therapy-based treatments, albeit with a lot of gaps that needs to be improved upon as a continuous process based on usage feedbacks and other indicators. However, we did not try this ourselves as that would require requesting paid access on their website.
    
    \item \textbf{Wysa} is an App-based interactive Chatbot for Anxiety Therapy (\url{https://www.wysa.com/}). It is rated 4.7 stars on Google Play store and is also available on Apple store. The AI agent, Wysa, provides interactive content that are tailored to users for boosting mental health. In the UK, the organisation works with the NHS by supporting ``patients through the NHS pathway starting from prevention support in the community, through interactive e-triage , waitlist support, AI guided CBT practice and relapse prevention.''(\url{https://www.wysa.com/nhs}). Their case study findings are summarised as: up to 80\% employees in a workplace ``saw their self-resilience improve after talking to Wysa for 8 weeks'', up to 91\% employees ``rated their conversations with Wysa as positive / helpful'' (\url{https://www.wysa.com/case-studies-and-reports}). 
    
    Similar to Tess, clinical evidences of Wysa's impact are reported in several researches; six of these are mentioned on their website (\url{https://www.wysa.com/clinical-evidence}), out of which, only one study did not have any authors affiliated to Wysa~\cite{leo2022}, and tested the efficacy of the Wysa app on reducing symptoms of depression and anxiety in a cohort of orthopedic patients with chronic musculoskeletal pain. Participants were adults between the age of 18 and 83 years, was undergoing treatment for orthopedic pain at the time of recruitment, and scored $\geq 55$ on the Patient-Reported Outcomes Measurement Information System (PROMIS), indicating presence of depression and anxiety in these patients. There were a total of 61 participants with a median age of 55 years and 87\% female. Over a period of 2 months of usage of the Wysa app, as well as to a new feature on the app specifically developed for users with chronic pain. The retention rate was 84\%; further, patients who completed the study were classified into `high' and `low' users of the app using statistical analysis. They report a significant decrease in anxiety scores among high users of the Wysa app, as well as a clinically meaningful decrease in depression scores (see figures 3~--~6 in~\cite{leo2022}). 
    
    Previously,~\cite{inkster2018} reported Wysa as an `empathy-driven' chatbot and tested its feasibility in improving mental well-being in a cohort recruited globally and who self reported depression using the PHQ-9 score. Here too, users were pooled into two groups viz.\ `high' and `low' users for analysing the data collected from participants over a period of approximately 8 weeks. Overall, the study reports an improvement in mood and depression among high user participants. The authors note that future work should record more longitudinal data points. 
    
    In the remaining four publications that are highlighted on the company website (all of which were reported during 2022), the research primarily focussed on understanding engagement of the users with the Wysa app; their interest group were patients with chronic pain who are often affected by mental health deterioration. The collective findings of these four researches provided directions for future improvements. Based on the publications and other resources available via their website, we speculate that the Wysa app is in the post-deployment testing phase. Their trial outcomes show promise and needs further evidence of efficacy and patient safety via longitudinal studies.

    \item \textbf{Wisdo} is available via the Google Play and Apple stores. Their website homepage carries the slogan `AI-Driven Peer Support Community' (\url{https://wisdo.com/}). Indeed, the keyword `loneliness' stands out in the showcased applications on their Resource page (\url{https://resources.wisdo.com/}), a rising trend especially among the elderly, but also among the young, including the workplace. The app relies on community support, including from people who have experienced similar circumstances to the user. In that regard, the app tries to personalise individual support in terms of both approach and content. Statistics demonstrating impact of the Wisdo app is reported under the `Results' tab of their webpage, claiming upto 60\% reduction in loneliness, 37\% reduction in depression and 39\% increase in overall improved health among users; also, using the app contributed to a reduction of around 1900 US Dollars per year in medial expenses for individual users. Furthermore, user retention is reported as 82\% over a period of 12 months. Reports on the app are disseminated via blogs and press releases, and can be accessed via their website.
     
     We have downloaded and tested the app on android. It provides a questionnaire targeting social media presence, availability of someone to talk to who has either gone through in the past, or going through, similar mental health issues as that of the user. Medical support is provided by humans, and not any AI agent. Thus, it is not clear as to how is the design of the App based on AI tools. The Wisdo App is rated 3.5 stars on the Google Play store.
\end{itemize}

We would also like to note an iOS App-based Chatbot `\textit{Shim}', that was developed by a Swedish company `Hello Shim' and has since been acquired by Kry (\url{https://nordic9.com/news/kry-acquired-hello-shim-a-swedish-chatbot-provider-for-improving-mental-well-being-news0874638850/}). Previously,~\cite{ly2017} had tested the efficacy of Shim on a cohort of 28 participants in the age range 26--49 years and randomised in a 1:1 ratio. They reported 78.6\% participants to be active at least 50\% of the days. Furthermore, feedback from participants on the content and functionalities of the app that provided some useful data on user's perspective and requirement from Chatbots in general. In addition, the responses indicated a tendency to develop a `bond' with the bot as a `medium' which are said to be supported by literature. Overall, Shim was reported to have `performed well' in comparison to Woebot (see Sec~\ref{sec:31}). Current status of Shim is not available to the best of our knowledge. Also, we did not find any reference to its underlying technology i.e.\ whether AI was used in developing the bot.  

Below, we discuss Chatbots that are proposed for Physical health and well-being.
\section{Chatbots for general assistive healthcare}
\label{sec:4}
Of the numerous endeavours or propositions of Chatbots for general medical healthcare, we discuss a selected few below to highlight the different areas where Chatbots are deemed to be useful viz. pediatric, pre-conception care, oncology, old age.

The ECA Gabby was designed by~\cite{gardiner2013} to assist women of the minority communities in the USA on `preconception care' (PCC), including mitigating risk factors prior to pregnancy, addressing issues like family planning, and such, towards a healthier pregnancy and post-natal phases. In that sense, Gabby acted as a `Virtual Patient Advocate' (VPA). Pilot tests showed that an average of 23 health risks were identified by each participant out of which they discussed 11 risks on the average with Gabby. Overall, Gabby was deemed as a promising means to mitigate the risk factors for preconception care.

`Pharmabot' is proposed as a human machine interface in pediatric medicine by~\cite{benilda2015}. The Chatbot providing information about general health problems in children and prescribing medicines. With the objective to complement the role of a pharmacist for pediatric medicine, Pharmabot was developed using Visual C\# and Microsoft Access as the front- and back-ends respectively, running on a stand-alone computer. Pharmabot is reportedly capable of prescribing generic medicines, along with information on dosage, possible side effects, and other drug related `small prints'. For this, a database is used. The authors reported statistical tests where responses from Pharmabot were validated with those from students and professionals in the field of medicine i.e. the domain experts.

~\cite{bickmore2013} have designed, developed and piloted an ECA over a period of two months on a culturally mixed cohort (122 participants, 21-69 years age range) in the USA to study effectiveness of digital interventions on improving lifestyle (physical activity and diet). The ECA is called by the name of `Karen' and the software needs installation and runs on a stand-alone computer. The ECA's knowledge-base (i.e.\ the database of possible questions and corresponding answers) is carefully designed by the authors based on the goals of this digital intervention. Two specific themes are followed by the ECA, viz. ACT and DIET, where the ECA assumes the role of a health counseller to improve pysical activity and fruit and vegetable consumption respectively. Their pre-pilot hypothesis was a positive impact of both ACT and DIET interventions. Physical activity was measured using the International physical activity questionnaire (IPAQ); fruit and vegetable consumption was measured using the ``NIH/NCI Fruit and Vegetable Scan (FVS)'' (see Sec. 2.1. in~\cite{bickmore2013} for further details on FVS). Their study showed a strong support for one of their four hypotheses; the DIET intervention scheme showed a statistically significant increase in FVS for all participants who completed the study (93\% of the participants). Overall, the study demonstrated that integrating digital agents for counselling in lifestyle changes is a promising trend, in spite its challenges in deployment for public health services.

In a series of research,~\cite{owens2015, owens2019} look into integrating ECA for informed decision making (IDM) in prostrate cancer treatment. Their research is focussed on African American (AA) males who are disproportionately prone to prostrate cancer. However, decision making about available treatment options is not straightforward; towards this,~\cite{owens2015} propose the use of `computer-based decision aids' (CBDAs) that ``include increased knowledge, enhanced informed decision-making self-efficacy, and reduced decisional conflict regarding cancer screening and/or treatment''. They propose dissemination of information to make informed choices through ECAs to patients. They introduce `iDecide', an ECA running on a stand-alone computer; algorithmic details are not specified, however, based on the descriptions and demonstrations, we speculate that it is rule-based. It had two sections, one of which was mainly educational about prostrate cancer, and the other engaged patients to prepare them for IDM~\cite{owens2019}. Their pilot study on AA males from South Carolina, USA, showed statistically significant improvement in prostrate cancer knowledge, although IDM post-intervention was not tested. In what seems like an evaluative study of using a `smartphone Chatbot' for supporting older patients undergoing chemotherapy as a cancer treatment at home,~\cite{piau2019} report that use of Chatbots in this regard is feasible. The use of Chatbot is reported to have saved 6 hours per nurse-time per patient. They concluded that in their `first phase of deployment', they found that Chatbots are a feasible option in their set objectives, and expressed a positive note on integration and acceptance by the local medical community in the near future. The only note on the exact technology used is that ``the application was a semi-automatised messaging system running on smartphones". Indeed, as (relatively, in context) early as 2016, a clinical trial conducted for cancer research at the Boston Medical Center, USA, had already implemented an ECA, `Study Buddy', as an `ECA Oncology Trial Advisor'~\cite{studybuddy2016}. Study Buddy was used for advising patients ``on chemotherapy regimens, promoting protocol adherence and retention, providing anticipatory guidance and answering questions, serving as a conduit to capture information about complaints or adverse events, providing a venue for communicating about updates.'' It was web-based and could be accessed by patients anywhere, anytime. Particulars about the design and development of Study Buddy, and indeed its pilot procedures, are not specified on the website.

Efficacy of Chatbots to overcome loneliness as well as eating disorders is addressed by~\cite{kramer2022}. They called their `e-Health service' as PACO consisting of two ECAs. However, their study did not show any major effects on the two stated disorders by using PACO. Furthermore, not only did they receive lower than expected participation in response to their advertisements, older adults in general were found to be reluctant altogether to participate in e-Health care. These results provide valuable insight for Chatbot developers.

Basit Ali and colleagues proposes a generic Chatbot~\cite{ali2021} that can be used for domain specific applications. Our interest is whether this Chatbot can be applied for medical diagnosis, which is what we focussed on when we read their article. The proposed Chatbot using three layers of typical machine learning tools to segregate and classify sentences, domain specific or otherwise. The first layer is a support vector machine (SVM) that categorises the input text, and associates a label with each sentence, that indicates the `intent’ of the sentence. The second layer does a pattern matching using cosine similarity measure between user inputs and the stored dataset. The third layer is a codec; a Bi-Long Short Term Memory (LSTM) is used as an encoder, and another LSTM is used as a decoder. The authors created a dataset that contain around 0.136 million question-and-answer (Q\&A) pairs that is used to train their model. There were 27 intent labels. Qualitative testing was done alongside quantitative testing. However testing protocols are not elucidated. 

Besides the above, we found a plethora of articles that have used deep and shallow learning techniques to build Chatbots. For example, an AI-based Chatbot~\cite{khan2017} uses Decision Tree algorithm for disease detection using input text from patients and provides prescriptions. The authors refer to the Chatbot as `Smart Doctor'. There is no report on testing and analysis of this Chatbot though. Along similar lines, K-Nearest Neighbour (KNN) classifiers were used in a Chatbot to identify disease based on user input of symptoms~\cite{mathew2019}. The testing of the framework is not clear, and the authors provide only snapshots of bot responses that matched their expectation. Another NLP-based Chatbot is proposed for similarity query-based assistance to patients by discussing the ailment and prescribing medicine~\cite{mendapara2021}. A research published during covid 2019 presents a query-based Chatbot~\cite{athota2020} that acts as a domain expert and provides answers to any queries that a patient may have. Users need to log in to the system with a username and password, which they can set during initial registration that in turn required users to share their key identification information for example gender, age, and contact details. The paper does not discuss matters on user data protection. Another effort to provide Chatbot assistance for covid related cases propose a Amazon Web Service (AWS)-based application using Amazon Lex for the bot design and AWS S3 services for data storage~\cite{das2022}. The authors refer it as Co-bot and display a set of simple queries related to Covid diagnosis. Snapshots of responses are provided as demonstration of the Co-bot response, and upto 70\% accuracy on the trained bot responses is reported. Furthermore, the co-bot is shown to supply general information on Covid 19. As the application is hosted by AWS, we presume that the bot is intended to be accessible on any digital device. The authors do not talk about the testing rigour and methodologies. ~\cite{cuayahuit2019} suggest that Reinforcement learning to train a Chatbot ensemble may provide greater flexibility in terms of the capability to provide medical assistance for wider variety of health conditions. In summary, we find that while there are indeed AI-based Chatbots for generic medicine and treatment, little is known about their advancement to deployment in healthcare presently.

\section{Discussion}
\label{sec:5}
Point-of-contact automation as an assistive tool in healthcare in the form of conversational agents is an attractive proposition. In this brief review, we have used the term Chatbots to represent machines that can communicate interactively and thus act as conversational agents. Our focus is on Chatbots that are enabled by Artificial Intelligence (AI). One of the primary challenges in the successful deployment of AI-enabled Chatbots in healthcare currently is a lack of trust and reliability on the technology. The first and foremost concern with machine-dependent healthcare is patient safety, followed by patient data confidentiality. It is thus not surprising that experts are now emphasising on `Trustworthy' and `Ethical' AI. There are primarily two modes of incorporating AI in Chatbots: (a) Supervised learning, where the knowledge-base of the machine is a finite universe, and it is trained to respond appropriately within this universe; (b) Unsupervised learning, where the machine learns by itself to respond to unknowns within a finite universe when trained with `experiential' examples from that universe. Needless to mention, hybrid modes (a combination of supervised and unsupervised) can also be used, such as generative AI eg.\ ChatGPT. Indeed, the advent of ChatGPT has set a high benchmark for natural language processing (NLP) skills of machines. However,as of now, its general responses cannot be trusted with making decisions and taking actions without evaluation by a human.

In our review, we have come across two types of (AI-enabled) Chatbots in healthcare; some that are designed for mental (eg. depression, anxiety) healthcare, and others that are intended for physical (eg. diabetes, cancer) healthcare. Of these, the Chatbots for mental healthcare are comparatively at an advanced stage of development. The ones that are currently advertised as `deployed' are commercial, and available at a cost to individuals and organisations; we have highlighted three such organisations that claimed using AI for developing the software for their respective Chatbots. The non-commercial Chatbots are led by not-for-profit-organisations viz.\ Higher Education, Research and Medical Institutions, and are in several stages of the Chatbot software development life-cycle. The most advanced stages that we have come across for these non-commercial Chatbots for mental healthcare is that of clinical trial for pilot tests before deployment. Furthermore, the most promising (as indicated on their websites) are the Chatbots MIRA (Canada) and Chatpal (Norther Periphery and Arctic Programme). Besides these, Woebot, a Chatbot for youth (college going) mental health, was used for clinical trials by several research groups outside of the proposing organisation; however, we did not find any references to current or future plans for Woebot. Several interesting Chatbots are proposed for general healthcare (eg Pharmabot for Pediatrics, Gabby for gynaecology, iDecide for oncology), many of which were pilot tested with clinical trials. We did not find any references to their current status or future plans for integration with public health services.

Our study indicates that there are huge opportunities for AI-enabled Chatbots in healthcare in both public and private sectors (eg. see \cite{druid2023}). In spite, there is a lack of trust in the technology among both the healthcare professionals and patients. To address this issue, an important step would be to initiate widespread collaboration and co-ordination between the medical and technology professionals to improve the Chatbot technology based on user/patient evaluation and feedbacks. Furthermore, rigorous evaluation via clinical trials is essential to establish reliability in terms of patient safety and data protection. From the patients' perspective, one way to build-up trust gradually on Chatbots could be by involving medical professionals `human-machine-interaction-loop'. Other than trust, some studies reported that patients of mental health expressed frustration and dissatisfaction over the reduced NLP skills of the Chatbot. With the arrival of ChatGPT, that may not be a cause for concern in times to come; using `narrow AI' and/or `prompt engineering' tactics, ChatGPT may well be trained to cater to bespoke healthcare services. While a lot has been achieved on the AI-enabled Chatbot technology, a lot still needs to be done for widespread integration with existing healthcare systems.

Being a brief and focussed review, we did not study several other non-AI Chatbots in healthcare (eg.\ Amaha, Sanvello) that score well in terms of user reviews Google Play Store. Furthermore, we do not claim to have covered all AI-enabled Chatbots, and neither is it feasible for this brief report in the interest of space. A wider review of Chatbots in healthcare, both in public and private sectors, that are not based on AI technology would be an interesting area to consider as future work.  

The recent global pandemic has emphasised the need for integration of assistive digital machine intelligence in the healthcare services. Such assistance from intelligent human-machine interactive technology (AI-enabled Chatbots) will only make healthcare systems stronger and better prepared for the future.

%\clearpage

\nolinenumbers

%This is where your bibliography is generated. Make sure that your .bib file is actually called library.bib
\bibliography{manuscript}

%This defines the bibliographies style. Search online for a list of available styles.
\bibliographystyle{abbrv}

\end{document}